\documentclass{bmvc2k}

\title{Zero-Shot Object Detection by Hybrid Region Embedding}

\addauthor{Berkan Demirel}{berkan.demirel@metu.edu.tr}{1}
\addauthor{Ramazan Gokberk Cinbis}{gcinbis@ceng.metu.edu.tr}{2}
\addauthor{Nazli Ikizler-Cinbis}{nazli@cs.hacettepe.edu.tr}{3}

\addinstitution{
 HAVELSAN Inc.\\
 Ankara, Turkey
}
\addinstitution{
 Middle East Technical University\\
 Ankara, Turkey
}
\addinstitution{
 Hacettepe University\\
 Ankara, Turkey
}

\runninghead{Demirel \textit{et al.}}{Zero-Shot Object Detection by Hybrid Region Embedding}

\def\eg{\emph{e.g}\bmvaOneDot}

 \usepackage{multirow}
 \input{macros}

\begin{document}

\maketitle

\begin{abstract}
Object detection is considered as one of the most challenging problems in computer vision, since it requires correct prediction of both classes and locations of objects in images. In this study, we define a more difficult scenario, namely \textit{zero-shot object detection (ZSD)} where no visual training data is available for some of the target object classes. We present a novel approach to tackle this ZSD problem, where a convex combination of embeddings are used in conjunction with a detection framework. For evaluation of ZSD methods, we propose a simple dataset constructed from Fashion-MNIST images and also a custom zero-shot split for the Pascal VOC detection challenge. The experimental results suggest that our method yields promising results for ZSD.
\end{abstract}

\section{Introduction}
\label{sec:intro}
Object detection is one of the most studied tasks in computer vision research. Previously, mainstream approaches 
provided only limited success despite the efforts in carefully crafting representations for object detection, \eg \cite{yan2014fastest}.
More recently, however, convolutional neural network (ConvNet) based models have lead to great advances in detection speed and accuracy,
\eg \cite{ren2015faster, redmon2016yolo9000, liu2016ssd}.

While the state-of-the-art in object detection is undoubtedly impressive, object detectors still lack semantic
scalability. As these approaches rely heavily on fully supervised training schemes,
one needs to collect large amounts of images with bounding box annotations for each target class of interest.
Due to its laborious nature, data annotation remains as a major bottleneck in semantically enriching and universalizing object detectors. 

Zero-shot learning (ZSL) aims to minimize the annotation requirements by enabling recognition of unseen classes, \ie those
with no training examples.  This is achieved by transferring knowledge from seen to unseen
classes by means of auxiliary data, typically obtained easily from textual sources.  Mainstream examples for such ZSL approaches include methods for
mapping visual and textual information into a joint space~\cite{akata2013label, akata2015evaluation, ba2015predicting},
and, those that explicitly leverage text-driven similarities across classes \cite{norouzi2013zero}. 

The existing ZSL approaches, however, predominantly focus on classification problems. 
In this work, we extend this ZSL paradigm to object detection and focus on the \textit{zero-shot
detection} (ZSD) task. Here, the goal is to recognize and localize instances of object classes with no training examples, purely based on 
auxiliary information that describes the class characteristics. 
The main motivation for studying ZSD is the observation that in most applications of zero-shot learning, such as robotics,
accurate object localization is equally important as recognition.

Our ZSD approach builds on the adaptation and combination of two mainstream approaches in zero-shot image
classification: (i) convex combination of class embeddings~\cite{norouzi2013zero}, and, label embedding based
classification~\cite{weston2010large}.  More specifically, we propose a hybrid model that consists of two components:
the first component leverages the detection scores of a supervised object detector to embed image regions into a class embedding space. 
The second component, on the other hand, learns a direct mapping from region pixels to the
space of class embeddings. Both of these region embeddings are then converted into region detection scores by comparing their similarities with true class embeddings.  
Finally, we construct our zero-shot detector by integrating these two components into the
the fast object detection framework YOLO~\cite{redmon2016yolo9000}.

We note that both components of our approach essentially provide an embedding of a given test image. Our main motivation
in using them together is to employ two complementary sources of information. In particular,  while the former component
can be interpreted as a semantic composition method guided by class detection scores, the latter one focuses on transformation of image content into the
class embedding space. Therefore, these two components are expected to better utilize semantic relations and visual cues, respectively.

In order to evaluate the effectiveness of the proposed ZSD approach, we create new benchmarks based on
existing datasets. First, we create a simple ZSD dataset by composing images with multiple
Fashion-MNIST~\cite{xiao2017fashion} objects. Moreover, the Pascal VOC~\cite{pascal_voc_IJCV} dataset is similarly
adapted to the ZSD task by defining new splits and settings. The experimental results demonstrate that our hybrid
embedding approach yields promising results in both datasets.

To sum up, our main contributions in this work are as follows: (i) we define a novel zero-shot setting for detecting
objects of unseen classes, (ii) we propose a novel hybrid method to handle newly defined ZSD task, (iii) we introduce 
two new benchmarks for evaluating ZSD approaches based on Fashion-MNIST and VOC datasets.

\begin{figure*}
\begin{center}
\includegraphics[width=\textwidth]{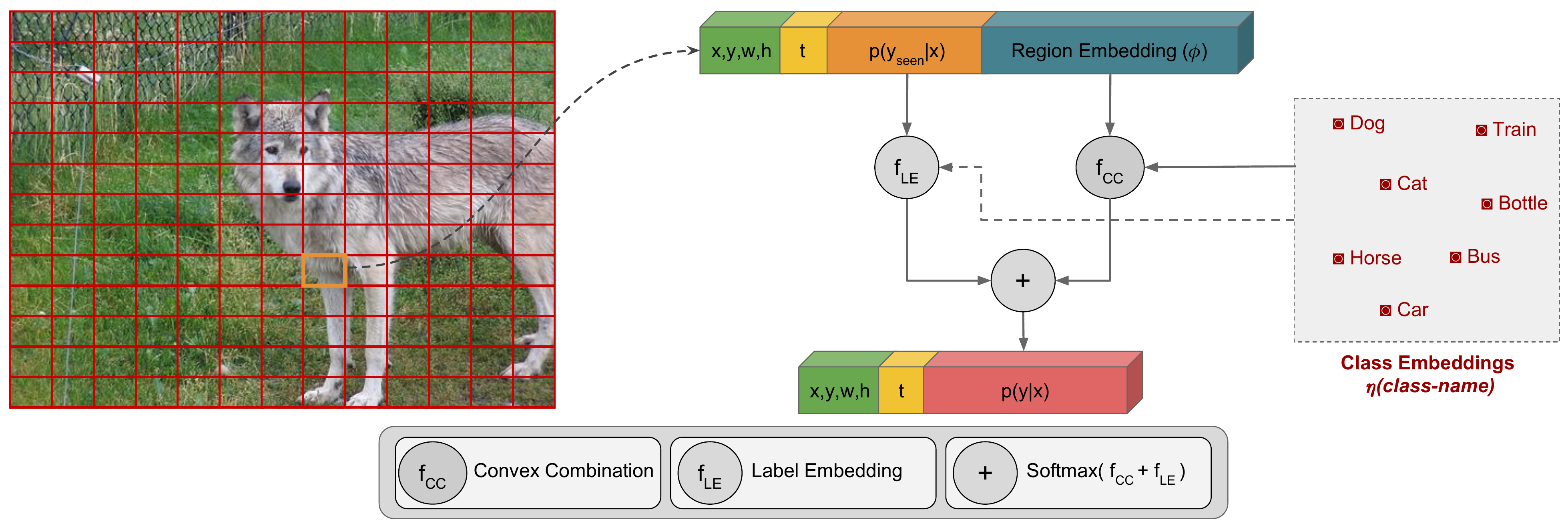}
\end{center}
    \caption{The framework of our ZSD model. In this model, $(x,y,h,w)$ represents bounding box regression coordinates, $t$ represents bounding box confidence score, $p(y_\text{seen}|c)$ represents initial class scores, $\phi$ represents embedding vector of related region, and $p(y|x)$ represents the final zero-shot detection class probabilities. }
\label{fig:main}
\end{figure*}

\section{Related Work}

In this section, we review related work on object detection and zero-shot learning.

\vspace{2mm}

\noindent\textbf{Object Detection.} State-of-the-art techniques in object detection are now all based on ConvNets. One
of the first ConvNet based object detection approaches is the OverFeat framework~\cite{sermanet2014overfeat}. OverFeat
is essentially a fully-convolutional network that jointly makes predictions for object classification, localization and
detection.
Girshick \textit{et al.}~\cite{girshick2014rich} propose R-CNN that extracts ConvNet features of pre-extracted object proposals. He
\textit{et al.}~\cite{he2015spatial} and Girshick \textit{et al.}~\cite{girshick2015fast} propose similar techniques
that allow efficient feature extraction for object proposals,
which greatly increases computational efficiency of the R-CNN-like approaches. Ren \textit{et al.}~\cite{ren2015faster}
propose a further improved architecture that jointly learns to generate and score object proposals.

Redmon \textit{et al.}~\cite{redmon2016you, redmon2016yolo9000} propose YOLO (You Only Look Once) architecture that 
also simulatenously generate candidate detection windows and per-window class probabilities. However, unlike 
R-CNN-like approaches, YOLO aims to produce all detections via a single feed-forward pass of a ConvNet, avoiding need to process
candidate windows in a sequential manner.
\textit{et al.}~\cite{liu2016ssd} use similar
idea via multi-scale feature maps, and yield detection results using convolutional filters. Additional approaches in
this context include using contextual and multi-scale representations~\cite{bell2016inside}, iterative grid without
object proposals~\cite{najibi2016g}, using feature pyramids~\cite{lin2017feature}, or, landmark
based detection~\cite{huang2015densebox}.

\vspace{2mm}
\noindent\textbf{Zero-Shot Learning.} Zero-shot learning is the process of transferring recognition knowledge from seen classes to unseen classes. Most ZSL approaches use prior information from some sources to encode and transfer necessary knowledge. Attributes are one of the most important source of prior information~\cite{lampert09,lampert13pami, luo2018zero, guo2018zero, demirel2017attributes2classname}. Using attributes at different abstraction levels and propagating on different levels provide more distinctive information for unseen classes~\cite{al2015transfer}. Other zero-shot learning methods are proposed, but in essence they commonly handle attributes, classes, features and their relations. Example studies use semantic class taxonomies~\cite{rohrbach2011evaluating}, hierarchy and exclusion (HEX) graphs~\cite{deng2014large}, random forests~\cite{jayaraman2014zero}, linear layered networks~\cite{romera2015embarrassingly}, semantic autoencoders~\cite{kodirov2017semantic}, visually meaningful word vectors~\cite{demirel2017attributes2classname}, semantic dictionary~\cite{ding2017low} or manifold regularizations~\cite{xu2017matrix}.

In recent years, label embedding methods have attracted attention in the zero-shot learning field. In this context, Akata \textit{et al.}~\cite{akata2013label,akata2015evaluation} modify WSABIE formula~\cite{weston2010large} and use distributed word representation as side information. Frome \textit{et al.}~\cite{frome2013devise} use convolutional neural network architectures for mapping visual features into a rich semantic embedding space. Song \textit{et al.}~\cite{song2018transductive} propose a transductive learning (QFSL) method to learn unbiased embedding space since embedding spaces often have strong bias problem. Besides, visual-semantic discrepancy problem is observed when using textual side information. In this context, Demirel \textit{et al.}~\cite{demirel2017attributes2classname} use attribute information as an intermediate layer to learn more generalizable distributed word representations.

Norouzi \textit{et al.}~\cite{norouzi2013zero} use convex combination of the semantic embedding vectors directly without learning any semantic space. Elhoseiny \textit{et al.}~\cite{elhoseiny2013write} handle zero-shot learning problem with purely textual descriptions. They define a constrained optimization formula that combine regression and knowledge transfer functions with additional constraints. Ba \textit{et al.}~\cite{ba2015predicting} use MLP in their text pipeline to learn classifier weights of CNN in the image pipeline to handle zero-shot fine-grained object classification. The defined MLP network generates a list of pseudo-attributes for each visual category by utilizing raw texts acquired from Wikipedia articles.

Finally, we note that while there are methods~\cite{redmon2016yolo9000, hoffman2014lsda, hoffman2015detector} for constructing object detectors through classifier-to-detector transfer, they still require carefully selected and annotated
training images to train source classification models. This is fundamentally different from the ZSD task, which requires
zero examples for unseen classes. By definition, therefore, ZSD is much more scalable towards large-scale
recognition, compared to classifier-to-detector transfer approaches.

\vspace{2mm}
\noindent\textbf{Recent works on ZSD.} We note that unpublished manuscripts of three independent works on zero-shot detection have very recently appeared on arXiv~\cite{bansal2018zero,zhu2018zero,rahman2018zero}. Despite the common theme of zero-shot detection, these works significantly differ from our hybrid model in the following ways: \cite{bansal2018zero} focuses on developing background-aware models,
\cite{rahman2018zero} proposes a semantic clustering loss, and \cite{zhu2018zero} uses attribute based models.
In addition, the experimental setup and datasets differ in all these works. We plan to investigate similarities and differences across these recent works and ours in future work.

\section{Method}

Our method consists of two components that (i) utilize a convex combination of class embeddings, an adaptation of the
ideas from \cite{norouzi2013zero}, and, (ii) directly learn to map regions to the space of class embeddings, by extending the
label embedding approaches from zero-shot image classification~\cite{akata2015evaluation}. 

The rest of this section
explains the model details: in the first two sub-sections, we describe the convex combination and label embedding
components. Then, we describe how we construct our
zero-shot object detector within the YOLO detection framework.

\subsection{Region Scoring by Convex Combination of Class Embeddings}

\def\fcc{f_\text{CC}} 
\def\ecc{\phi_\text{CC}} 
\def\Y{\mathcal{Y}}
\def\Ys{\mathcal{Y}_s}
\def\Yu{\mathcal{Y}_u}
\def\Ns{|\mathcal{Y}_s|}
\def\Nu{|\mathcal{Y}_u|}
\def\fle{f_\text{LE}}   
\def\ele{\phi_\text{LE}} 
\def\cls{\eta} 
\def\Lle{L_\text{LE}}

First component of our ZSD approach aims to semantically represent an
image in the space of word vectors. More specifically, we represent a given image region (\ie a bounding box)
as the convex combination of training class embeddings, weighted by the class scores given by a 
supervised object detector of seen classes. The resulting semantic representation of the region is then utilized to
estimate confidence scores for unseen classes.

This approach can be specified as follows: let $\Ys$ be the set of seen classes,
for which we have training images with bounding box annotations, and and let $\Yu$ be the set of unseen classes,
for which we have no visual training examples. Our goal is to learn a scoring function 
$\fcc(x,b,y): \mathcal{X}\times\mathcal{B}\times\mathcal{Y}\rightarrow\mathcal{R}$ that measures the relevance of label $y \in \Ys$, which can be a seen or unseen class, for a given candidate bounding box $b \in \mathcal{B}$ and the image $x \in \mathcal{X}$. 

We assume that a $d_e$ dimensional embedding vector $\cls(y)$, such as word embeddings of class names or class-wise attribute
indicator vectors, is available for each class.  The scoring function $\fcc(x,b,y)$ is then defined as the cosine
similarity between the class embedding $\cls(y)$ and the image region embedding $\ecc(x,b)$:
\begin{equation}
\fcc(x,b,y) = \frac{\ecc(x,b)^\text{T} \cls(y)}{\|\ecc(x,b)\| \|\cls(y)\|}
\label{eq:func_fcc}
\end{equation}
where $\ecc(x,b)$ is defined as follows:
\begin{equation}
\ecc(x,b) = \frac{1}{\sum_{y \in \Ys} p(y|x,b)} \sum_{y \in \Ys} p(y|x,b) \cls(y)
\label{eq:conv}
\end{equation}
Here, $p(y|x,b)$ is the class posterior probability given by the supervised object detection model. Therefore,
$\ecc$ can simply be interpreted as a weighted sum of class embeddings, over the seen classes.

\subsection{Region Scoring by Label Embedding}

The convex combination driven scoring function $fcc$ utilizes detection scores and embeddings of the 
training classes to estimate scores of zero-shot classes. In the label embedding approach, however, 
our goal is to directly model the compatibility between the visual features of image regions
and class embeddings.
For this purpose, we define the label embedding driven scoring function $\fle(x,b,y): \mathcal{X}\times\mathcal{B}\times\mathcal{Y}\rightarrow\mathcal{R}$
that measures the relevance of label $y \in \Y$ for a given candidate bounding box $b \in \mathcal{B}$ in an image $x \in \mathcal{X}$ as follows:
\begin{equation}
\fle(x,b,y) = \frac{\ele(x,b)^\text{T} \cls(y)}{\|\ele(x,b)\| \|\cls(y)\|}
\label{eq:func_fle}
\end{equation}
where $\ele(x,b)$ is basically a deep convolutional neural network that maps the 
image region $b$ of image $x$ to the space of class embeddings.

We note that $\fle(x,b,y)$ can equivalently be interpreted as a dot product between $\ell_2$-normalized
image region descriptors and class embeddings. While 
it is common $\ell_2$-normalize class embeddings in zero-shot image classification studies \cite{akata2015evaluation}, we also $\ell_2$-normalize the image embedding vectors. In our preliminary experiments, we have observed that this additional normalization step is beneficial for the zero-shot detection task. 

We learn the $\ele(x,b)$ network in an end-to-end fashion within our YOLO-based zero-shot detection framework, which we
explain in the next section.

\subsection{Zero-Shot Object Detection}

We use the YOLO-v2~\cite{redmon2016yolo9000} architecture to construct our zero-shot object detector. The original YOLO architecture
that we utilize contains a convolutional network that reduces the spatial dimensions of the input by a factor of 32 and results in a tensor of depth $k(5+\Ns)$,
\eg an input image of size $416 \times 416 \times 3$ results in a tensor of size $13 \times 13 \times k(5+\Ns)$.
Each cell within this output tensor encodes the $k$ detections per cell ($k=5$ by default), and, each block of size $5+\Ns$ encodes one such detection. Here, for a single detection, the first 4 dimensions encode the relative bounding box coordinates, the following dimension encodes the estimated window objectness score,
and the final $\Ns$ dimensions encode class confidence scores.

To adapt YOLO architecture for the zero-shot detection task, we modify it in the following manner: we increase the final output depth 
from $k(5+\Ns)$ to $k(5+\Ns+d_e)$, where the newly added $d_e$ dimensions per detection correspond to the $\ele(x,b)$ output of the label embedding component of the model.
In this way, the same convolutional network is shared for candidate box prediction, class prediction and class-embedding prediction purposes.

During training, the original YOLO formulation uses three separate mean-squared error based loss functions, defined over
the differences between predictions and ground truth values for (i) bounding boxes, (ii) intersection-over-union values, and, (iii) 
classes. For training $\fcc$ defined in \Eq{func_fcc}, the original YOLO loss function over class predictions is used as is.
For training $\fle$ defined in \Eq{func_fle}, however, we extend the loss function by incorporating an additional loss function $\Lle$.
$\Lle$ basically measures correctness of the label embedding driven class predictions in a max-margin sense:
\begin{equation}
%
    %\Lle(x,b,y) = \frac{1}{\Ns - 1} \sum_{y^\prime \neq y, y^\prime \in \Ys} \max\left(0,1-\fle(x,b,y)+\fle(x,b,y^\prime)\right)
    \Lle(x,b,y) = \frac{1}{\Ns - 1} \sum_{y^\prime \in \Ys \setminus \{y\}} \max\left(0,1-\fle(x,b,y)+\fle(x,b,y^\prime)\right)
\end{equation}

where $y$ is the ground-truth class corresponding to the bounding box $b$ in input image $x$. Here, the goal is to ensure that 
at each window prediction, the label embedding based confidence score $\fle$ for the target class is larger than that
of each other class. Other than this extension, we use the original YOLO training procedure, over the seen classes.

Once the network is trained, we jointly utilize the scoring functions $\fcc$ and $\fle$ by computing soft-max of their summations, over the classes of interest:

\begin{equation}
    p(y|x,b) = \frac{\exp\left(\fcc(x,b,y)+\fle(x,b,y)\right)}{\sum_{y^\prime \in \Y}\exp\left(\fcc(x,b,y^\prime)+\fle(x,b,y^\prime)\right)}
\label{eq:output_f}
\end{equation}
where $p(y|x,b)$ is the predicted posterior probability of (seen or unseen) class $y$ given region $b$ of image $x$. The final set of detections are obtained by using the non-maxima suppression procedure of YOLO over all candidate detection windows, objectness scores, and the final probabilities $p(y|x,b)$.

\section{Experiments}

In this section, we present our experimental evaluation of the proposed approach.  In \sect{dataset}, we describe the
ZSD datasets that we prepare and utilize.  In \sect{classembed}, we explain class embeddings used in our experiments.
Finally, in \sect{fashionexp} and \sect{pascalexp}, we give the implementation details and our experimental results.

\subsection{Datasets}
\label{sec:dataset}

We use two different datasets: Fashion-ZSD and Pascal-ZSD. 
We propose two new testbeds for evaluation of ZSD approaches. First, we create a synthetic dataset based on combinations
of objects from the Fashion-MNIST~\cite{xiao2017fashion} dataset. Second, 
we compose a new split based on existing Pascal VOC~\cite{pascal_voc_IJCV} benchmarks.
The details of these testbeds are described below.

\vspace{2mm}

\begin{figure}
\resizebox{\textwidth}{!}{%
\begin{tabular}{cccc}
\includegraphics[width=2.9cm]{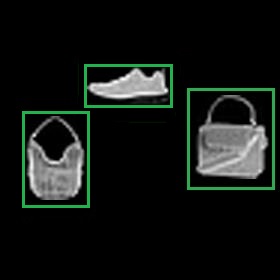}&
\includegraphics[width=2.9cm]{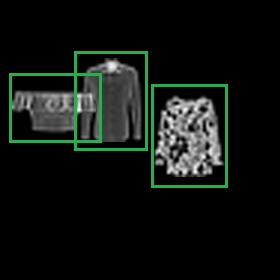}&
\includegraphics[width=2.9cm]{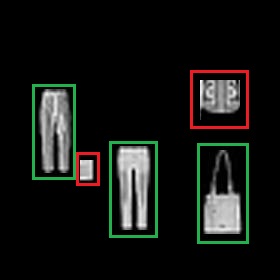}&
\includegraphics[width=2.9cm]{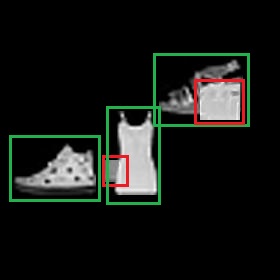}\\
(a)&(b)&(c)&(d)
\end{tabular}
}
\caption{Ground truth object regions are shown with green and noise regions are shown in red boxes. The dataset consists of images from four different scenarios. From left-to-right, (a) full objects only, (b) partial occlusions, (c) clutter regions included, and (d) a scene with both partial occlusions and clutter regions.}
\label{fig:fashion_dataset}
\end{figure}

\noindent \textbf{Fashion-ZSD:} This is a toy dataset that we generate for evaluation of ZSD methods, based on the Fashion-MNIST~\cite{xiao2017fashion} dataset. Fashion-MNIST originally consists of Zalando's article images with associated labels. This dataset contains 70,000 grayscale images of size 28x28, and 10 classes. For ZSD task, we split the dataset into two disjoint sets; seven classes are used in training and three classes are used as the unseen test classes (Table~\ref{table:zsd_fashion_results}). We generate multi-object images such that there are three different objects in each image. Randomly cropped objects are utilized to create clutter regions. As shown in Figure~\ref{fig:fashion_dataset}, we consider four scenarios: no noise or occlusion, scenes with partial occlusions, those with clutter, and, finally scenes with both partial occlusions and clutter regions. $8000$ images of the resulting $16333$ training images are held out for validation purposes. As a result, we obtain the Fashion-ZSD dataset with 8333 training, 8000 validation and 6999 test images. 

\vspace{2mm}

\noindent \textbf{Pascal-ZSD.} This is an adapted version of the Pascal VOC datasets~\cite{pascal_voc_IJCV}. We select 16 of the 20 classes for training and the remaining 4 classes (\ie car, dog, sofa and train) for test. The {\em train+val} subsets of Pascal VOC 2007 and 2012 datasets are used for training classes, and the {\em test} subset of Pascal 2007 is used for evaluation on the unseen classes. Images containing a mixture of train and test classes are ignored. 

\subsection{Class Embeddings}
\label{sec:classembed}

For the Fashion-ZSD dataset, we generate 300-dimensional GloVe word embedding vectors~\cite{pennington2014glove} for each class name, using Common Crawl Data\footnote{commoncrawl.org/the-data/}. For the class names that contain multiple words, we take the average of the word vectors. For Pascal-ZSD, we use attribute annotations of aPaY dataset~\cite{farhadi2009describing}, since aPascal(aP) part of this dataset is obtained from Pascal VOC images. We 
average 64-dimensional indicator vectors of per-object attributes over the dataset to obtain class embeddings. 

\subsection{Zero-Shot Detection on Fashion-ZSD Dataset}
\label{sec:fashionexp}

In this part, we explain our ZSD experiments on Fashion-ZSD dataset. We initialize the convolutional layers of our model
using the weights pre-trained on the ILSRVC12\cite{deng2009imagenet} classification images. Training of our approach is completed in 10 epochs,
where batch size is 32 and learning rate is 0.001. In our experiments, we first evaluate the performance of the trained
network on seen training classes. According to the results presented in Table~\ref{table:zsd_fashion_results}, the
proposed approach obtains $91.9\%$ mAP on the validation images with seen classes, which shows the proper training
of the detection model. On the test set with unseen classes only, our proposed approach yields an mAP of
$64.9\%$, highlighting the difficulty of zero-shot detection task even in simple settings.

On the combinated validation and test evaluation, our method achieves $81.7\%$ mAP. This setting
is particularly interesting, as it requires recognition over both seen and unseen objects at detection time.
Our result suggests that the model is able to detect objects of unseen test classes even in the presence of seen 
classes, without being dominated by them.

\begin{table}[]
\centering
\resizebox{\columnwidth}{!}{
\begin{tabular}{c|ccccccc|ccc|c}
         & \multicolumn{7}{c|}{Training Classes}                       & \multicolumn{3}{c|}{Test Classes} &          \\
Test split  & t-shirt & trouser & coat & sandal & shirt & sneaker & bag  & pullover  & dress  & ankle-boot  & mAP (\%) \\
\hline
val      & .89    & .91    & .90  & .97   & .86  & .99    & .90  & -         & -      & -           & 91.9    \\
test     & -       & -       & -    & -      & -     & -       & -    & .49      & .49   & .95        & 64.9    \\
val+test & .89    & .90     & .90  & .97   & .86  & .99    & .91 & .45      & .40    & .90         & 81.7  
\end{tabular}
}
\caption{ZSD performances of proposed hybrid method on Fashion-ZSD dataset. We report class based average precision and mean average precision (mAP) scores.}
\label{table:zsd_fashion_results}
\end{table}

\subsection{Zero-Shot Detection on Pascal-ZSD Dataset}
\label{sec:pascalexp}

In this part, we explain our ZSD experiments on Pascal-ZSD dataset. Training settings of the proposed method on
Pascal-ZSD dataset are same with the previous experiment, except that the number of epochs is set to 30. 
We present the results our approach, as well as individual performances of convex combination and label embedding components,
in Table~\ref{table:zsd_pascal_Results}. The proposed hybrid approach yields $65.6\%$ mAP on seen
classes, $54.6\%$ mAP on unseen classes and $52.3\%$ mAP on the combination of 
seen and unseen classes. By comparing individual components of the model, we observe that 
convex combination (CC) outperforms label embedding (LE), and the hybrid scheme further improves the results.

\vspace{4mm}

\begin{figure}
\resizebox{\textwidth}{!}{%
\begin{tabular}{ccc}
\includegraphics[width=4cm,height=3cm]{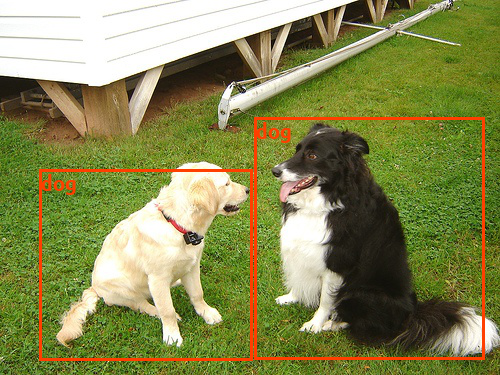}&
\includegraphics[width=4cm,height=3cm]{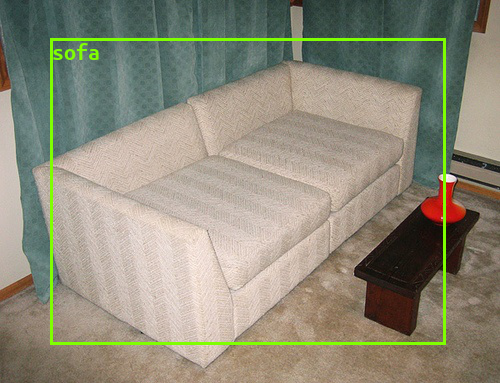}&
\includegraphics[width=4cm,height=3cm]{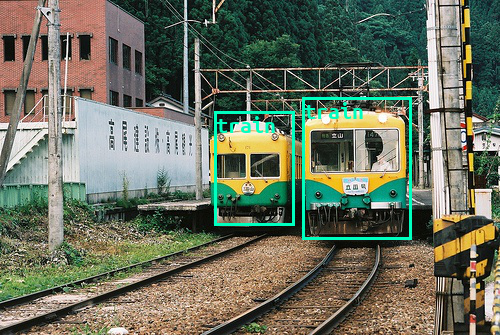}\\
\includegraphics[width=4cm,height=3cm]{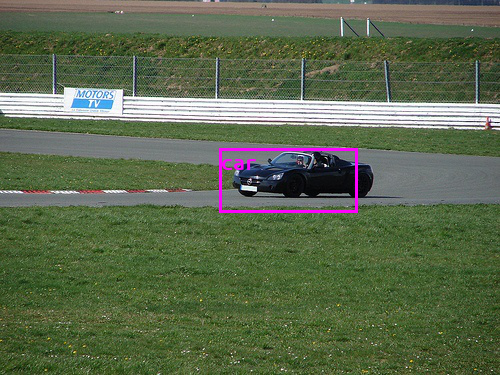}&
\includegraphics[width=4cm,height=3cm]{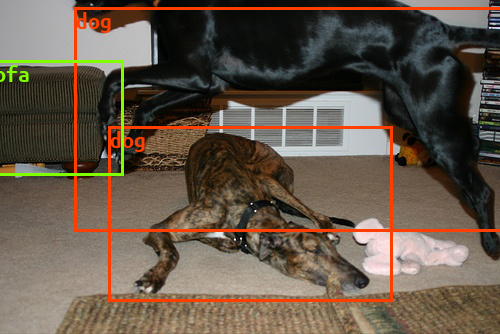}&
\includegraphics[width=4cm,height=3cm]{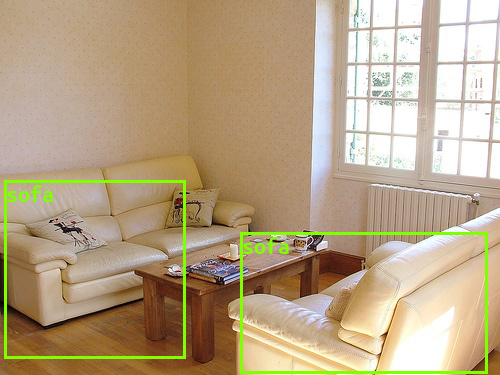}\\
\includegraphics[width=4cm,height=3cm]{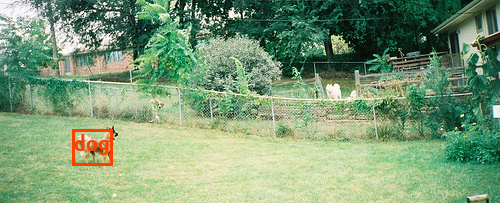}&
\includegraphics[width=4cm,height=3cm]{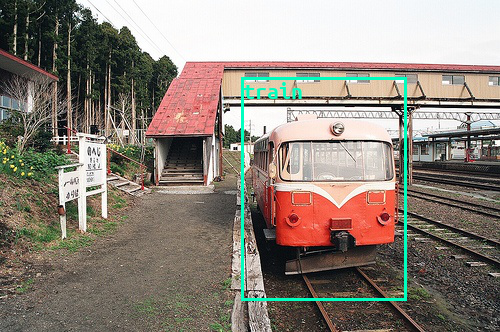}&
\includegraphics[width=4cm,height=3cm]{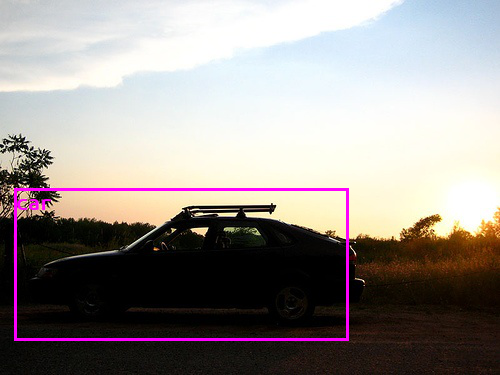}\\
\end{tabular}
}
\caption{Successful detection results of unseen objects on Pascal-ZSD dataset using proposed hybrid region embedding.}
\label{fig:pascal_results}
\end{figure}

The reason why the performance of the individual label embedding component is much lower can potentially be explained by the
fact that the ZSD-Pascal dataset is relatively small: there are 16 classes in the training set, and this number is
most probably insufficient to learn a direct mapping from visual features to class embeddings.

\begin{table}[]
\centering
\resizebox{\columnwidth}{!}{%
\begin{tabular}{c | c | cccccccccccccccc|cccc|c}
\rotatebox{90}{Method}             & \rotatebox{90}{Test split}  & \rotatebox{90}{aeroplane} & \rotatebox{90}{bicycle} & \rotatebox{90}{bird} & \rotatebox{90}{boat}  & \rotatebox{90}{bottle}  & \rotatebox{90}{bus}   & \rotatebox{90}{cat}  & \rotatebox{90}{chair}  & \rotatebox{90}{cow}  & \rotatebox{90}{dining table} & \rotatebox{90}{horse} & \rotatebox{90}{motorbike} & \rotatebox{90}{person} & \rotatebox{90}{potted plant} & \rotatebox{90}{sheep}  & \rotatebox{90}{tvmonitor}   & \rotatebox{90}{car}  & \rotatebox{90}{dog}  & \rotatebox{90}{sofa} & \rotatebox{90}{train} & \rotatebox{90}{mAP (\%)}       \\
\hline
\multirow{3}{*}{LE} & v      & .46 & .50  & .44 & .28 & .12 & .59 & .44 & .20  & .11 & .38 & .35 & .47 & .65 & .16 & .18 & .53 & -    & -    & -    & -    & 36.8          \\
                   & t     & -    & -    & -    & -    & -    & -    & -    & -    & -    & -    & -    & -    & -    & -    & -    & -    & .54 & .79 & .45 & .12 & 47.9          \\
                   & v+t & .34 & .48 & .40  & .23 & .12 & .34 & .28 & .12 & .09 & .32 & .28 & .36 & .60  & .15 & .13 & .50  & .27 & .26 & .20  & .05 & 27.4          \\
\hline
\multirow{3}{*}{CC} & v      & .69 & .74 & .72 & .63 & .43 & .83 & .73 & .43 & .43 & .66 & .78 & .80  & .75 & .41 & .62 & .75 & -    & -    & -    & -    & 65.0          \\
                   & t     & -    & -    & -    & -    & -    & -    & -    & -    & -    & -    & -    & -    & -    & -    & -    & -    & .60  & .85 & .44 & .27 & 53.8          \\
                   & v+t & .67 & .73 & .70  & .59 & .41 & .61 & .58 & .32 & .32 & .65 & .74 & .68 & .72 & .39 & .57 & .72 & .49 & .24 & .10  & .15 & 52.0          \\
\hline
\multirow{3}{*}{H} & v      & .70  & .73 & .76 & .54 & .42 & .86 & .64 & .40  & .54 & .75 & .80  & .80  & .75 & .34 & .69 & .79 & -    & -    & -    & -    & \textbf{65.6} \\
                   & t     & -    & -    & -    & -    & -    & -    & -    & -    & -    & -    & -    & -    & -    & -    & -    & -    & .55 & .82 & .55 & .26 & \textbf{54.2} \\
                   & v+t & .68 & .72 & .74 & .48 & .41 & .61 & .48 & .25 & .48 & .73 & .75 & .71 & .73 & .33 & .59 & .57 & .44 & .25 & .18 & .15 & \textbf{52.3}
\end{tabular}
}
\caption{ZSD performances of proposed label embedding (LE), convex combination (CC) and hybrid (H) methods on Pascal-ZSD dataset. We report class based average precision and mean average precision (mAP) scores of validation (v), test (t) and validation+test (v+t) images.}
\label{table:zsd_pascal_Results}
\end{table}

\begin{figure}
\resizebox{\textwidth}{!}{
\begin{tabular}{ccc}
\includegraphics[width=6cm,height=4cm]{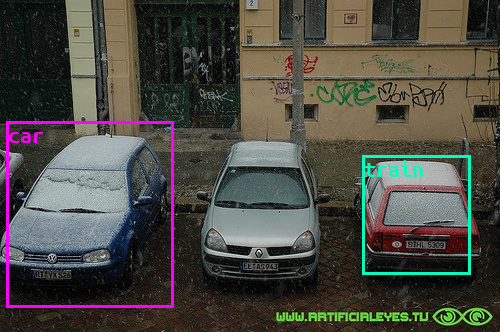}&
\includegraphics[width=6cm,height=4cm]{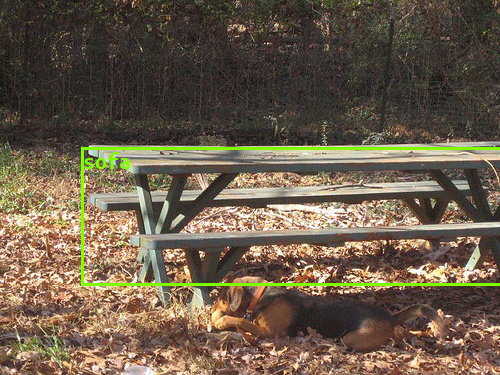}&
\includegraphics[width=6cm,height=4cm]{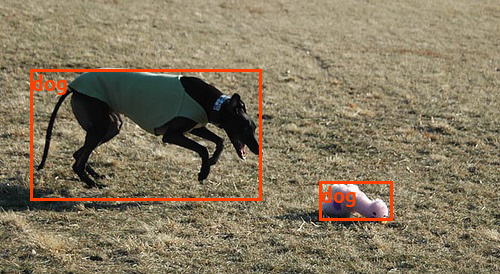}\\
\end{tabular}
}
\caption{Unsuccessful detection results of unseen objects on Pascal-ZSD dataset using hybrid region embedding.}
\label{fig:pascal_results_failure}
\end{figure}

Qualitative results for our approach are provided in Figure~\ref{fig:pascal_results}. In this figure, example results of succesful detections
of objects of unseen classes with various poses and sizes are shown.
Additionally, example failure cases are shown on Figure~\ref{fig:pascal_results_failure}. Problems in detection include
missed detections, false positives, as well as misclassification of objects despite correct localization.
For instance, in the second image within
Figure~\ref{fig:pascal_results_failure}, we see that \textit{''picnic bench''} object is misrecognized  as \textit{''sofa''},
most probably due to relative similarity of the '\textit{'chair''} and \textit{''dining table''} seen classes in the embedding space.

\section{Conclusion}
Accurate localization of unseen classes is equally important as recognition of them in various applications such as robotics. Moreover, to overcome the bottleneck of annotation, better ways of enriching object detectors are needed. To this end, in this work, we handle the problem of zero-shot detection and propose a novel hybrid method that aggregates both label embeddings and convex combinations of semantic embeddings together in a region embedding framework. By integrating these two components within an object detector backbone, detection of classes with no visual examples becomes possible. We introduce two new testbeds for evaluating ZSD approaches, and our experimental results indicate that the proposed hybrid framework is a promising step towards achieving ZSD goals.

\section{Acknowledgements}
This work was supported in part by the TUBITAK Grant 116E445.

\bibliography{egbib}
\end{document}